\documentclass[letterpaper, 10 pt, conference]{ieeeconf}  

\IEEEoverridecommandlockouts                              

\usepackage{graphics} 
\usepackage{epsfig} 
\usepackage{mathptmx} 
\usepackage{multirow}
\usepackage{booktabs}
\usepackage[table,xcdraw]{xcolor}
\usepackage[normalem]{ulem}
\useunder{\uline}{\ul}{}
\usepackage{hyperref}
\usepackage{color, colortbl}
\usepackage{stfloats}
\usepackage{float}
\usepackage{amsfonts} 

\let\labelindent\relax
\usepackage{enumitem}

\newcommand\egno{\textit{e.g.}}
\newcommand\ieno{\textit{i.e.}}
\usepackage{amsmath} 
\title{\LARGE \bf
Self-evolved Imitation Learning in Simulated World
}

\author{Yifan Ye$^{1}$, Jun Cen$^{2}$, Jing Chen$^{3,4,\dagger}$, Zhihe Lu$^{1,\dagger}$
\thanks{Yifan Ye and Zhihe Lu are with $^{1}$College of Science and Engineering, Hamad Bin Khalifa University, Education City, Doha 24404, Qatar}%
\thanks{Jun Cen is with $^{2}$College of Computer Science and Technology, Zhejiang University, Hangzhou 310058, China}%
\thanks{
Jing Chen is with $^{3}$School of Physics and Optoelectric Engineering, and $^{4}$Guangdong Provincial Key Laboratory of Sensing Physics and System Integration Applications, Guangdong University of Technology, Guangzhou 510006, China
}
\thanks{$^{\dagger}$ Corresponding Authors, jchen125@gdut.edu.cn and zlu@hbku.edu.qa}
}

\begin{document}

\maketitle
\thispagestyle{empty}
\pagestyle{empty}

\begin{abstract}

Imitation learning has been a trend recently, yet training a generalist agent across multiple tasks still requires large-scale expert demonstrations, which are costly and labor-intensive to collect.
To address the challenge of limited supervision, we propose Self-Evolved Imitation Learning (SEIL), a framework that progressively improves a few-shot model through simulator interactions. 
The model first attempts tasks in the simulator, from which successful trajectories are collected as new demonstrations for iterative refinement. 
To enhance the diversity of these demonstrations, SEIL employs dual-level augmentation: (i) Model-level, using an Exponential Moving Average (EMA) model to collaborate with the primary model, and (ii) Environment-level, introducing slight variations in initial object positions.
We further introduce a lightweight selector that filters complementary and informative trajectories from the generated pool to ensure demonstration quality.
These curated samples enable the model to achieve competitive performance with far fewer training examples. 
Extensive experiments on the LIBERO benchmark show that SEIL achieves a new state-of-the-art performance in few-shot imitation learning scenarios.
Code is available at \href{https://github.com/Jasper-aaa/SEIL.git}{https://github.com/Jasper-aaa/SEIL.git}.

\end{abstract}

\section{INTRODUCTION}

Imitation Learning (IL) \cite{haldar2024baku,Chi2023DiffusionPV,Florence2021ImplicitBC,Shafiullah2022BehaviorTC} has demonstrated remarkable success across a variety of tasks by leveraging large-scale expert demonstration datasets \cite{RT1}. 
However, collecting such datasets is often time-consuming, labor-intensive \cite{libero}, and in certain domains (\egno, surgical practice) can even be impractical \cite{Wah2025a}. 
This motivates the study of IL under limited supervision, which is both a more practical and challenging setting. 
However, learning from only a few demonstrations typically results in a substantial performance drop. For instance, in the one-shot setting, a train-from-scratch Diffusion Policy \cite{Chi2023DiffusionPV} achieves only 0.8\% success rate, compared to 50.5\% with full training.

\begin{figure}[ht]
    \vspace{-0.3cm}
    \setlength{\abovecaptionskip}{-0.6cm}
    \includegraphics[width=0.5\textwidth]{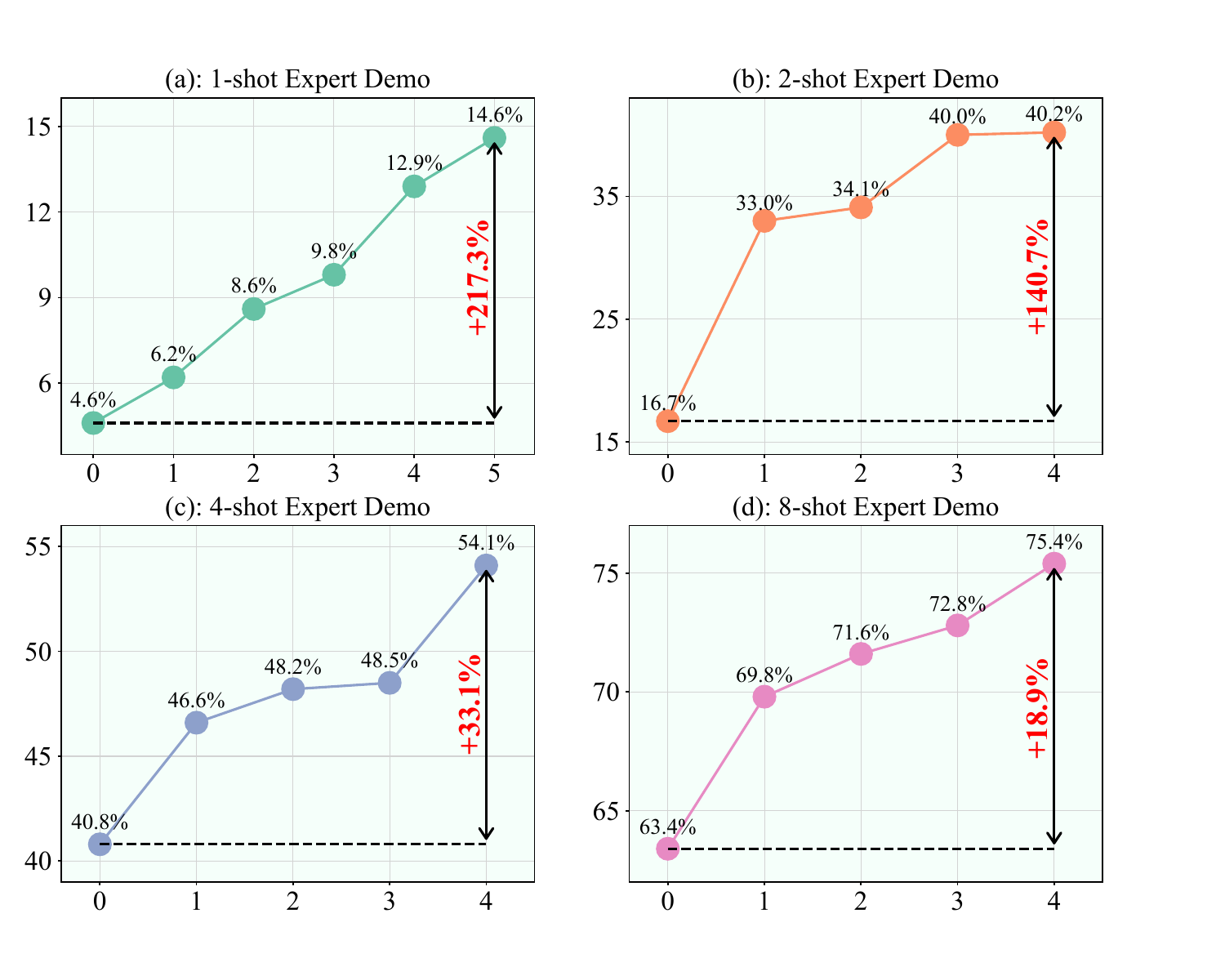}
    \caption{Performance of the proposed SEIL on Libero Long. The x-axis denotes the number of evolutionary stages, while the y-axis represents the success rate on the benchmark at each stage. We report the performance growth rates throughout the evolution process.}
    \label{fig:performance}
    \vspace{-0.5cm}
\end{figure}

Inspired by the empirical success of Reinforcement Learning (RL) \cite{Silver2016MasteringTG,Ibarz2021HowTT,Li2023MAHALOUO,Rashidinejad2021BridgingOR,Dey2023ReinforcementLB}, where agents explore through extensive trial-and-error within simulators, a natural solution is to leverage simulation environments to autonomously generate additional demonstrations for fine-tuning few-shot imitation policies.
That is, a policy is initially trained on a limited set of expert demonstrations and then deployed in a simulator, where it interacts with the environment and records successful trajectories based on simple success/failure feedback. 
Once a sufficient number of demonstrations across tasks are collected, the policy is refined using standard imitation learning updates on the newly acquired data. 
However, this approach gives rise to two critical challenges:
(i) How can we ensure sufficient diversity in the simulated demonstrations to support iterative learning? 
(ii) How can we identify and select the most informative samples to optimize model performance while maintaining computational efficiency?


In this paper, we investigate a few-shot imitation learning setting,  address the aforementioned two key challenges and propose Self-Evolved Imitation Learning (SEIL), a novel framework that leverages simulator to enable policy self-evolution from limited expert demonstrations, often with significant improvements as shown in Figure \ref{fig:performance}.
SEIL is built upon two core components: dual-level augmentation, comprising model-level and environment-level strategies, for generating diverse trajectories, and a sample selector tailored to identify and retain the most informative demonstrations for efficient policy refinement.
Notably, SEIL supports multiple rounds of interaction with the simulator, enabling the policy to evolve progressively over time.
Specifically, for model-level augmentation, SEIL introduces an auxiliary model to conduct additional rollouts alongside the primary model. 
To avoid incurring extra training overhead, particularly important in a multi-stage evolution setting where cost accumulates over iterations, this auxiliary policy is implemented as an Exponential Moving Average (EMA) of the main model. 
The EMA model maintains diversity in the policy space without the need for separate training, thereby supporting efficient and scalable demonstration generation.
In addition to model-level augmentation, we introduce environment-level augmentation, which enables the policy to interact with the simulator under diverse initial conditions. 
This variation in starting states enhances environmental diversity, thereby enriching the recorded demonstrations and improving the robustness of policy learning.

After generating a pool of diverse demonstrations via dual-level augmentation, we introduce a lightweight selector to identify informative samples.
The efficiency nature of this selector comes from taking as input only an initial image and a trajectory, rather than full video sequences.
Trained on few-shot data via a trajectory classification task, the selector learns category-specific feature patterns. 
Once trained, the selector is frozen and applied to all recorded demonstrations. 
Samples with the lowest confidence, \ieno, those most distinct from expert demonstrations, are selected for policy training.

Our contributions are summarized as follows.
\begin{itemize}[left=0pt]
    \item We investigate a few-shot imitation learning setting and propose Self-Evolved Imitation Learning (SEIL), a novel framework that leverages simulation environments to enable policy self-evolution from limited expert demonstrations.
    \item We propose dual-level augmentations, \ieno, model-level and environment-level augmentation, to guarantee the diversity of demonstration generation.
    \item We propose a lightweight selector to identify the most informative demonstrations, enhancing both policy performance and training efficiency. 
    \item Extensive experiments demonstrate that SEIL can effectively evolve weak few-shot trained models, achieving substantial improvements, \egno, a 217.3\% performance growth over the 1-shot baseline on Libero-Long.
\end{itemize}

\section{Related Work}
\subsection{Learning from Offline Data}
Imitation Learning (IL) \cite{Chi2023DiffusionPV} aims to train models to replicate expert behavior and is typically framed as a supervised learning problem. 
In contrast, Reinforcement Learning (RL) \cite{offline-dropo,offline-implicit,zhao2022offline} requires carefully designed, often complex, reward functions to guide policy learning. 
Recently, the integration of IL and RL has gained popularity \cite{SurveyDeepReinforcement,DRLsurvey,Kim2022DemoDICEOI}, where IL provides a foundational understanding of task dynamics from offline expert demonstrations, and RL augments the learned policy with reasoning and generalization capabilities. 
However, prior works \cite{Li2023MAHALOUO,Rashidinejad2021BridgingOR,Dey2023ReinforcementLB} largely emphasize the RL component.
In contrast, our proposed SEIL framework highlights the strengths of IL.
SEIL avoids the need for reward engineering by using demonstration-based supervision and is explicitly designed to achieve strong performance from limited expert data. 
It employs a gradual self-refinement strategy that leverages simulated feedback to evolve the policy iteratively. 
This design fundamentally differs from conventional offline RL approaches, which typically focus on leveraging large-scale static datasets in a single training phase \cite{Christiano2017DeepRL,agarwal2020optimistic,WhatMattersLearning}.


\subsection{Few-shot Imitation Leanrning}
Few-shot Imitation Learning (FIL) has been introduced to address scenarios where large-scale expert demonstrations are unavailable \cite{mete2024quest}. 
Some recent approaches leverage large language or vision foundation models \cite{chen2025vidbot,Stone2023OpenWorldOM}, which exhibit strong generalization to new domains with only a few examples. 
Others utilize generative models to augment expert demonstrations by increasing diversity in objects, backgrounds, and contexts \cite{yuScalingRobotLearning2023,lin2024data}. 
However, incorporating foundation models typically incurs additional computational costs, and generative augmentation can produce unrealistic or misleading trajectories. 
In this work, we propose the Self-Evolved Imitation Learning (SEIL) framework, which enables a weak, few-shot-initialized policy to gradually evolve. 
Instead of relying on generative models, SEIL leverages simulator-generated demonstrations, grounded in real interaction dynamics, to safely and effectively expand the training dataset without introducing inconsistencies.

\begin{figure*}[!ht]
    \centering
    \includegraphics[width=\textwidth]{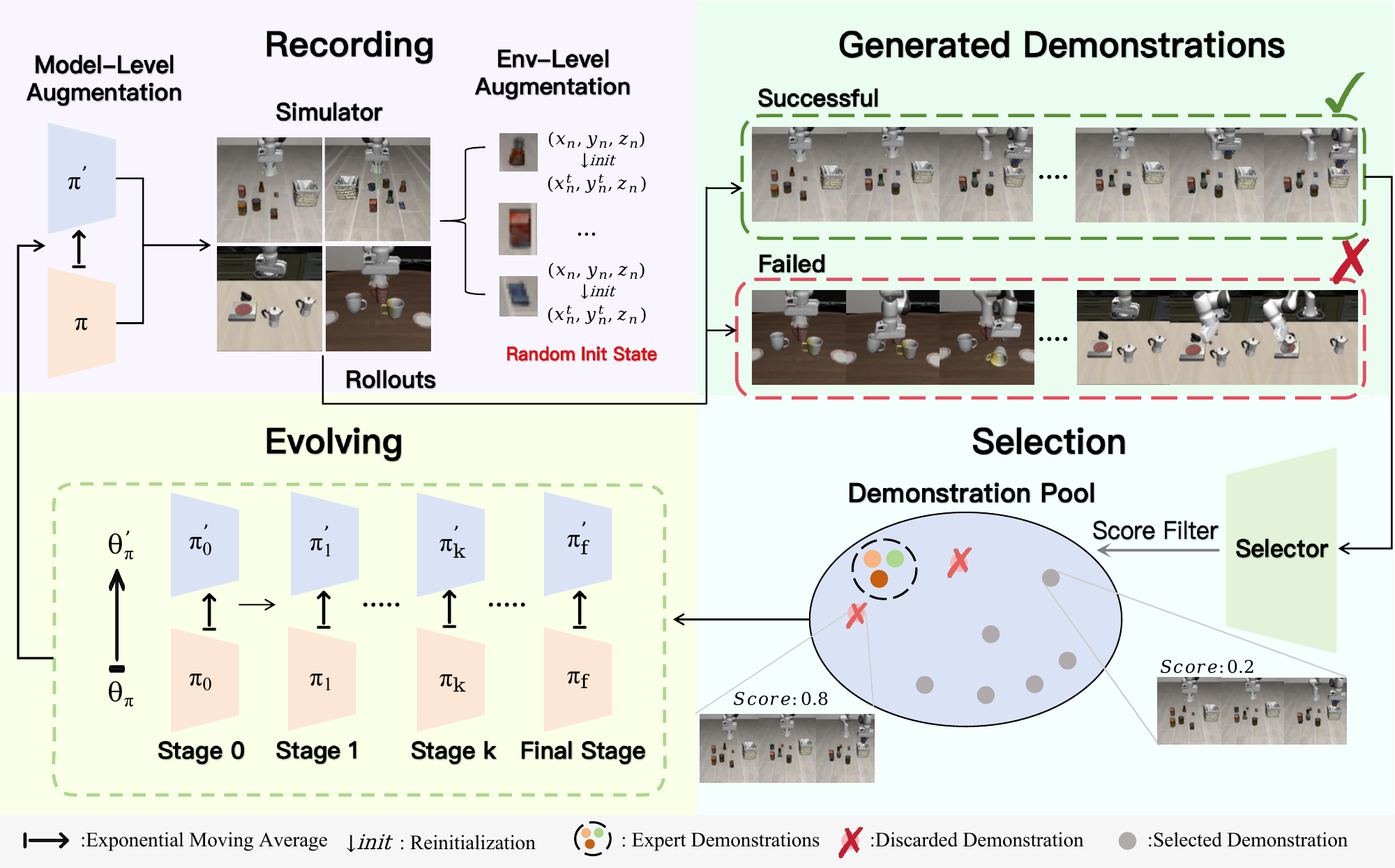}
    \setlength{\abovecaptionskip}{-0.3cm}
    \caption{Overall framework of our proposed SEIL Method. The framework begins by training a model $\pi_{0}$ using few-shot expert demonstrations. We then use the trained model to record demonstrations in the simulator. During this process, we apply model-level augmentation, where the trained policy $\pi_{0}$ interacts with the simulator alongside the EMA-enhanced model $\pi_{0}^{'}$, and environment-level augmentation, where the initial states are continuously varied to promote trajectory diversity. From simulator rollouts, we retain successful demonstrations and select the most informative using our proposed selector. The selected demonstrations incrementally expand the pool, allowing the model to evolve through iterative imitation learning updates until converging to the final policy $\pi_{f}$.}
    \vspace{-0.1cm}
    \label{fig:framework}
\end{figure*}

\section{Primaries}
\subsection{Problem Formulation}
In this paper, we aim to enhance the performance of a multi-task imitation learning (IL) model trained on few-shot expert demonstrations by leveraging a simulator $S$. 
We consider an expert dataset 
$D_{\text{expert}} = \{\{(O_{1}, A_{1}), \ldots, (O_{M_i}, A_{M_i})\}_{i=1}^{N_u}, T_u\}_{u=1}^{L}$, consisting of $L$ tasks, where $A_M$ denotes an action for task $T$ and $O_M$ is the corresponding robot observation. 
We focus on the low-data regime with $N_{u} \in \{1, 2, 4, 8\}$, consistent with standard few-shot settings \cite{yu2023task,li2024graphadapter}. 
The simulator $S$, composed of task $T$ and environment $E$, is used to generate the next observation $O_{t+1}$ based on the provided action $A_t$ at timestep $t$.

\subsection{Baseline Model}
We adopt BAKU \cite{haldar2024baku}, an efficient multi-task learner, as our baseline model. 
BAKU is composed of three key components: sensory encoders, an observation trunk, and an action head. The sensory encoders first process raw inputs from the environment and encode them into compact representations, which are then passed to the observation trunk to integrate information across different modalities. Finally, the action head predicts a sequence of actions based on the fused representation.

\section{Proposed Methods}
In this section, we describe the key components of the proposed SEIL framework.
Section \ref{meth:frame} outlines the overall architecture of SEIL, which is designed to progressively refine a few-shot model initialized from scratch.
Section \ref{meth:demo} introduces the dual-level augmentation strategies employed to ensure diversity in the generated demonstrations. 
Section \ref{meth:select} details the design and training of our lightweight selector for identifying informative demonstrations.

\subsection{SEIL Framework}\label{meth:frame}

To improve a weak policy $\pi_0$ trained on few-shot expert demonstrations, we enable it to interact with the simulator across multiple trials using a \textit{dual-level augmentation} strategy. 
Based on success/failure feedback from the simulator, we collect all successful rollouts, referred to as Recorded Demonstrations (RD). 
We then apply the proposed \textit{trajectory selector} to identify the most informative samples from RD, which are combined with expert demonstrations to retrain the policy. 
As illustrated in Figure \ref{fig:framework}, this iterative cycle, \textit{train–record–select–train}, constitutes the core of the Self-Evolved Imitation Learning (SEIL) framework.
The policy progressively evolves through $\pi_1, \pi_2, \dots, \pi_k$, until convergence is reached at $\pi_f$, where performance saturates and the self-evolution process concludes.

\subsection{Dual-Level Augmentation}\label{meth:demo}

To enrich the diversity of RD, we propose a simple yet effective dual-level augmentation strategy. 
The first component, Environment-Level Augmentation, randomizes the simulator’s initial states to expose the policy to varied environmental conditions. 
The second component, Model-Level Augmentation, leverages an auxiliary model with different parameters to generate complementary trajectories, enhancing behavioral diversity.


\subsubsection{Environment-Level Augmentation}
We first introduce Environment-Level Augmentation ($E_{\text{Aug}}$). 
As illustrated in the top-left panel of Figure \ref{fig:framework}, prior to each interaction between the policy and the simulator, the position of each object is randomly perturbed. 
This transformation is defined as:
\begin{equation}
(x, y, z) \xrightarrow{init} (x_{\text{new}}^{t}, y_{\text{new}}^{t}, z),
\label{eq:env}
\end{equation}
where $(x, y, z)$ denotes the default initial position of an object in the simulator, and $\xrightarrow{init}$ represents the random initialization function. 
A small offset is applied along the $x$ and $y$ axes to induce spatial variation while preserving vertical consistency.

\subsubsection{Model-Level Augmentation}
In Model-Level Augmentation ($M_{\text{Aug}}$), we introduce an auxiliary model to generate additional trajectories, with the key requirement that it incurs no extra training overhead. 
Since SEIL involves a multi-stage self-evolution process, training an additional policy at each stage would significantly increase the overall computational cost. 
To address this, we adopt an Exponential Moving Average (EMA) model, a lightweight, ensemble-style technique that avoids backpropagation while still producing diverse and stable rollouts for data augmentation.

Specifically, given the parameters at step $t$ of model $\pi$, denoted as $\theta_t$, the EMA parameters $\theta^{EMA}_t$ are updated as:
\begin{equation}
    \theta^{EMA}_t = \tau \cdot \theta^{EMA}_{t-1} + (1-\tau)\cdot \theta_t,
    \label{eq:ema}
\end{equation}
where $\tau \in [0,1)$ is the decay rate that controls the contribution of the most recent parameters. Intuitively, $\theta^{EMA}_t$ serves as a smoothed version of the online model parameters, thus providing better generalization and evaluation stability.

\subsubsection{Collaboration}
We deploy both the augmented model $\pi'$ and the base model $\pi$ in conjunction with the environment augmentation $E_{\text{Aug}}$, forming our Dual-Level Augmentation strategy. 
This combination effectively enhances the diversity of the generated trajectories. 
Implementing both $E_{\text{Aug}}$ and $M_{\text{Aug}}$ simultaneously is essential. 
If only $E_{\text{Aug}}$ is used, the evolution process relies solely on the base model $\pi$, resulting in slow progress due to limited behavioral variability. 
On the other hand, using $M_{\text{Aug}}$ alone introduces inconsistencies between input–action pairs, \ieno, $(O_t, A_t^{\text{Base}})$ versus $(O_t, A_t^{\text{EMA}})$, as the base and augmented models may produce divergent behaviors under identical observations. 
We further analyze the impact of these configurations in Section~\ref{EXP:comp}.

\subsection{Trajectory Selector}\label{meth:select}

\begin{figure}[!h]
    \centering
    \includegraphics[width=\linewidth]{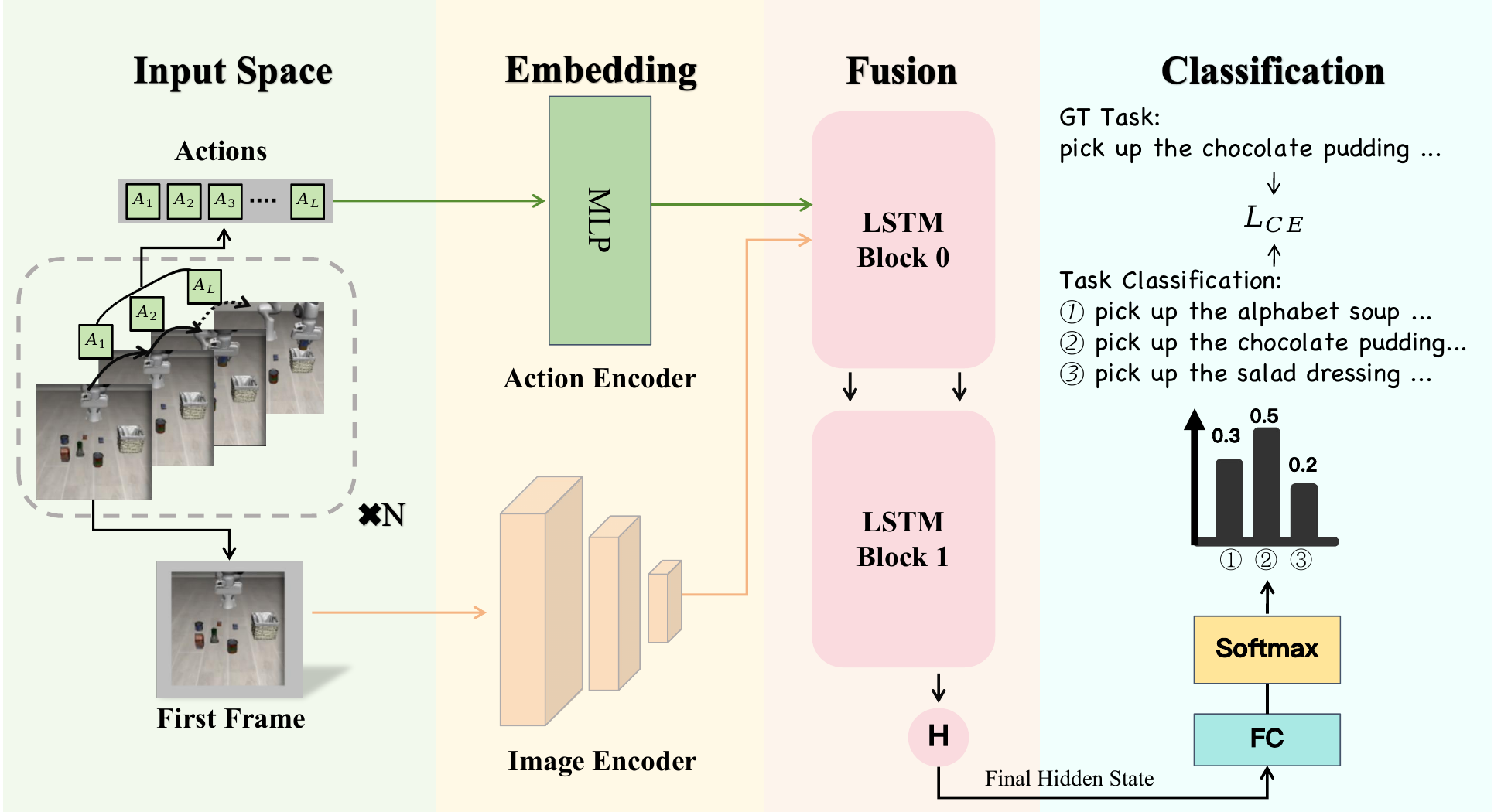}
    \setlength{\abovecaptionskip}{-0.3cm}
    \caption{
    Architecture of the proposed trajectory selector. We propose a lightweight trajectory selector composed of an image encoder and an action encoder, which process the first-frame image and the corresponding action sequence, respectively. The encoded features are fused and passed through an LSTM, followed by an MLP that outputs a confidence score for each trajectory.
    }
    \label{fig:selector}
\end{figure}

\subsubsection{Overview}

We further propose a lightweight selector to identify and retain the most informative samples from the demonstration pool for policy evolution.
Intuitively, effective selection requires the selector to capture both the distribution of RD and the action semantics within each trajectory. 
To this end, we train the selector using a task classification objective, enabling it to learn the underlying distribution of expert demonstrations. 
Once trained, the selector evaluates the RD and assigns confidence scores, allowing us to identify trajectories that are either representative or complementary to expert data, thereby enhancing diversity and effectiveness during model evolution.

\subsubsection{Architecture Design}

Inspired by the concept of world models \cite{cen2025worldvlaautoregressiveactionworld}, we adopt a similar formulation for demonstration selection. 
That is, a world model $\Phi_w$ predicts the next observation $O_{t+1}$ based on a history of $h$ observation-action pairs, formally expressed as:
\begin{equation}
O_{t+1} = \Phi_w(O_{t-h:t}, A_{t-h:t}).
\end{equation}

Motivated by this idea, we represent each demonstration using only its first frame and the corresponding action sequence, effectively decomposing the full trajectory into a compact representation. 
This design choice significantly reduces computational cost while preserving essential temporal and semantic information, thus enhancing the selector’s capability to assess and differentiate demonstrations.
Concretely, as illustrated in Figure \ref{fig:selector}, given an input image $\mathbf{x} \in \mathbb{R}^{C \times H \times W}$ and an action sequence $A$, we employ a ResNet-18\cite{resnet} as the image encoder $\Phi_i$ and a multi-layer perceptron (MLP) as the action encoder $\Phi_a$ to extract feature representations. 
The encoding process is formally defined as:
\begin{equation}
\begin{aligned}
\mathbf{f}_A = \Phi_a(A), \quad \mathbf{f}_{\mathbf{x}} = \Phi_i(\mathbf{x}),
\end{aligned}
\label{eq:encode}
\end{equation}
where $\mathbf{f}_A$ and $\mathbf{f}_{\mathbf{x}}$ denote the action and image embeddings, respectively.

We then concatenate the two embeddings to obtain a fused representation $\mathbf{f}_t = [\mathbf{f}_{\mathbf{x}}; \mathbf{f}_{A}]$, which is fed into a Long Short-Term Memory (LSTM) network to model temporal dependencies. 
The final hidden state $\mathbf{H}$ produced by the LSTM is passed through a classification MLP to generate task logits. The entire selector is optimized using the cross-entropy loss based on ground-truth task labels.

\paragraph{Discussion}

A natural approach to demonstration task classification is to either (i) train a video model to classify full-length demonstration videos, or (ii) use a sequence model to classify trajectories based solely on the action sequences. 
While both methods can learn the distribution of the RD, they come with notable drawbacks. 
Video models, though expressive, introduce significant computational overhead and are susceptible to overfitting, especially in few-shot settings with limited expert demonstrations. 
Sequence-only models, on the other hand, ignore visual context and thus struggle to capture the semantics of recorded demonstrations. 
In contrast, our design strikes a balance between efficiency and representational power by using the first-frame image and the associated action sequence as input. 
This hybrid representation captures both visual and temporal cues while maintaining low computational cost, enabling effective task classification without the drawbacks of the above alternatives.

\subsubsection{Selection Scheme}
Once the selector is trained, we leverage its confidence scores over the RD to guide the selection of diverse training samples. 
We explore and compare three sampling strategies:

\begin{itemize}[left=0pt]
\item \textit{Uniform sampling:} Randomly selecting demonstrations regardless of confidence, used as a baseline.
\item \textit{High-confidence sampling:} Selecting demonstrations with the highest confidence, \ieno, those most similar to expert data.
\item \textit{Low-confidence sampling:} Selecting demonstrations with the lowest confidence, representing those most distinct from expert data.
\end{itemize}

While high-confidence sampling prioritizes expert-like behavior, it limits exposure to diverse trajectories and thus restricts policy generalization. 
In contrast, low-confidence sampling emphasizes diversity by selecting demonstrations farthest from the expert distribution, leading to richer learning signals and improved policy refinement. 
Our experiments show that this strategy consistently yields superior performance compared to the other alternatives.

\section{Experiments}
\subsection{Evaluation Benchmark}

\subsubsection{Benchmark}
In our experiments, we adopt the LIBERO benchmark \cite{libero}, which includes four sub-categories: LIBERO-Spatial, LIBERO-Object, LIBERO-Goal, and LIBERO-Long. 
LIBERO-Spatial evaluates spatial reasoning by requiring the robot to position a bowl based on its current location. 
LIBERO-Object tests object recognition and manipulation of distinct items. 
LIBERO-Goal focuses on procedural generalization by varying task goals while keeping the object set fixed. 
LIBERO-Long consists of ten long-horizon tasks designed to assess the robot’s ability to perform extended and temporally complex operations.

\subsubsection{Baselines}
We compare our approach against four baselines: Action-Chunking Transformer (ACT) \cite{act}, Diffusion Policy (DP) \cite{Chi2023DiffusionPV}, RT-1 \cite{RT1}, and BAKU \cite{haldar2024baku}. 
ACT is a transformer-based architecture that predicts temporally chunked action sequences. 
DP frames action prediction as a denoising diffusion process. 
RT-1 is a large-scale vision-language-action model trained on a diverse set of real-world robotic tasks.

\subsubsection{Implementation Details}
We implement our method on top of BAKU \cite{haldar2024baku}, following all training settings. In evolving, we allow both models to perform 25 rollouts in the simulator each round and use the selector to choose 15 of them to incrementally augment the evolving demonstration pool. Empirically, we set the EMA decay parameter $\tau$ to 0.999.
For fairness, all the baselines and our method are trained from scratch. All ablation studies are conducted on the 8-shot Libero Long setting. 
For the selector, we use a pretrained ResNet-18 as the image encoder and stack two LSTM blocks, with an embedding hidden dimension of 256. 
The model is trained using Adam optimizer with a learning rate of 0.001 for 80 epochs. 

\subsubsection{Metrics}
For the policy evaluation, we follow the standard LIBERO evaluation, each task is evaluated for 50 rollouts under benchmark initial states and we record the success rate (SR). 
For the selector, we report the classification accuracy on a validation set, which is constructed from 20 randomly sampled unseen expert demonstrations in Libero.

\begin{table}[th]
\centering
\caption{Comparison on Libero Dataset.}
\label{tab:main_result}
\begin{tabular}{c|c|cccc}
\toprule[1.0pt] 
Dataset & Methods    & 1 shot & 2 shot & 4 shot & 8 shot \\ \hline 

\multirow{5}{*}{Libero Long}    
               & DP       &  0.8\%  & 6.4\%    & 9.2\%    & 11.2\%   \\
               & ACT        &  0.0\%  & 4.7\%   & 13.0\%   & 15.8\%  \\               
               & RT-1       &  0.0\%  & 0.0\%    & 0.0\%    & 0.0\%   \\
               & BAKU       & 4.6\%   & 16.7\%   & 40.8\%   & 63.4\%  \\
               & SEIL(Ours) & \textbf{14.6\%}  & \textbf{40.4\%}   & \textbf{54.3\%}   & \textbf{75.4\%}  \\ \hline

\multirow{5}{*}{Libero Spatial} 
               & DP       &  19.8\%  & 28.0\%    & 31.2\%    & 36.6\%   \\
               & ACT        &  8.0\%  & 17.8\%   & 48.4\%   & 50.4\%  \\               
               & RT-1       & 0.0\%   & 0.0\%    & 0.0\%    & 0.0\%  \\
               & BAKU       & 42.2\%  & 51.8\%  & 75.2\%  & 83.6\%  \\
               & SEIL(Ours) & \textbf{53.2\%}  & \textbf{71.4\%}  & \textbf{87.2\%}  & \textbf{87.7\%}  \\ \hline

\multirow{5}{*}{Libero Object}  
               & DP       &  28.4\%  & 32.8\%    & 43.4\%    & 65.6\%   \\
               & ACT        &  15.8\%  & 28.6\%   & 52.8\%   & 68.4\%  \\               
               & RT-1       & 0.0\%   & 0.1\%  & 0.1\%  & 0.1\%  \\
               & BAKU       & 65.0\%  & 65.1\%  & 69.8\%  & 96.8\%  \\
               & SEIL(Ours) & \textbf{82.0\%}  & \textbf{81.8\%}  & \textbf{95.2\%}  & \textbf{99.2\%}  \\ \hline

\multirow{5}{*}{Libero Goal}    
               & DP       &  20.0\%  & 33.8\%    & 39.2\%    & 59.0\%   \\
               & ACT        &  5.4\%  & 23.9\%   & 21.0\%   & 47.2\%  \\               
               & RT-1       & 0.0\%   & 12.7\%   & 15.2\%  & 18.2\%  \\
               & BAKU       & 30.3\%  & 48.0\%   & 71.6\%  & 84.8\%  \\
               & SEIL(Ours) & \textbf{49.2\%}  & \textbf{79.0\%}   & \textbf{84.2\%}  & \textbf{92.8\%}  \\ \hline

\multirow{5}{*}{Average} 
               & DP       &  17.2\%  & 25.2\%    & 30.7\%    & 43.1\%   \\
               & ACT        & 7.3 \%  & 18.8\%   & 33.8\%   & 45.4\%  \\                
                & RT-1       & 0.0\%  & 3.2\%  & 3.8\%  & 4.6\%  \\
                & BAKU       & 35.5\% & 45.4\% & 64.3\% & 82.2\% \\
                & SEIL(Ours) & \textbf{49.7\%} & \textbf{68.1\%} & \textbf{80.2\%} & \textbf{88.7\%} \\
\bottomrule[1.0pt]
\end{tabular}
\end{table}

\subsection{Main Results}

Table \ref{tab:main_result} shows that our proposed SEIL consistently outperforms all baselines across various few-shot settings, often with significant margins, highlighting its strong effectiveness in data-scarce scenarios. 
We further make three key observations.
First, strong VLA models such as DP, ACT, and RT-1 suffer severe performance degradation under limited supervision. 
For instance, in the 1-shot setting, their success rate fall below 1\%, while BAKU achieves a modest 4.6\%. 
In contrast, SEIL reaches 14.6\%, yielding a 217.3\% improvement over BAKU, demonstrating robustness in long-horizon tasks even with minimal demonstrations.
Second, SEIL achieves comparable or superior performance with fewer shots. For example, it obtains 87.2\% accuracy in the 4-shot setting on Libero-Spatial, outperforming BAKU’s 83.6\% in the 8-shot setting.
Finally, while all methods benefit from increased supervision, SEIL and BAKU exhibit notably stronger gains.
This suggests that both methods possess architectures that better leverage additional data, with BAKU’s design particularly suited for such scaling.

\begin{figure*}[h]
    \centering
    \includegraphics[width=\linewidth]{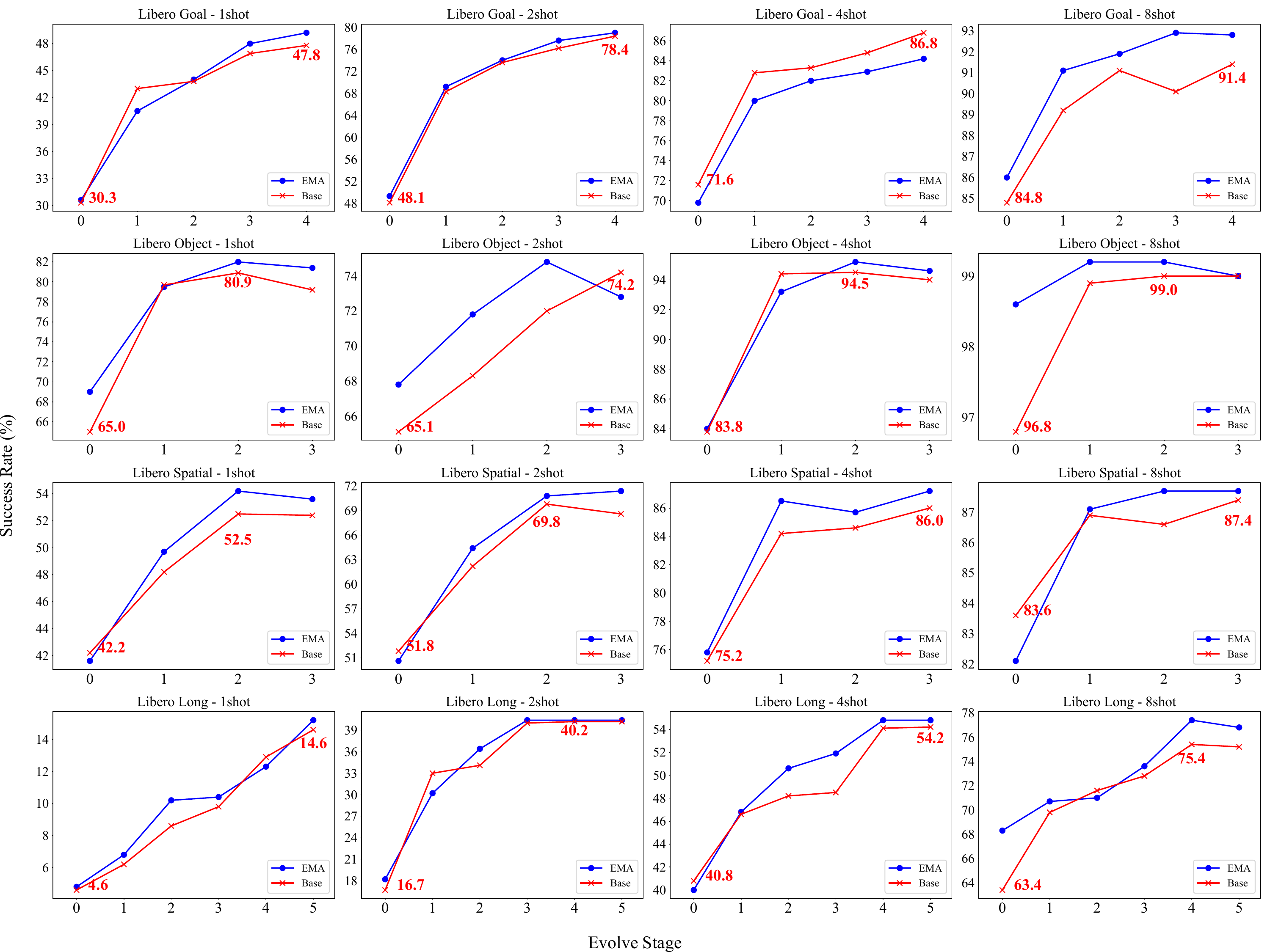}
    \caption{Success rate curves for different few-shot settings in the Libero Goal, Libero Object, Libero Spatial and Libero Long tasks.
The red and blue lines represent the success rate of the Base model and EMA model, respectively, across each round. Additionally, the red text indicates the success rate of the Base model prior to evolution (Baseline) and after evolution (Best Round).}
    \label{fig:4_data}
    \vspace{-0.4cm}
\end{figure*}

We further visualize SEIL’s evolution process in Figure \ref{fig:4_data}, where results indicate steady and consistent performance improvements across stages, ultimately converging once performance saturation is reached. Notably, our method typically yields a substantial improvement in the first round. For instance, in the 2-shot Libero Long setting, performance improves from 16.7\% to 33.0\%, and in the 4-shot Libero Goal setting, it improves from 71.6\% to 83.3\% in the first round.


\subsection{Ablation Study}

\begin{table}[t]
\caption{Ablation study on different components. We report the success rate after one round ($\text{SR}_{1}$) and four rounds ($\text{SR}_{4}$) of evolution. Here, $M_{Aug}$ and $E_{Aug}$ denote Model-Level Augmentation and Environment-Level Augmentation, respectively. ``-'' denotes no $E_{Aug}$ for the first round.}
\centering
\label{tab:components}
\begin{tabular}{ccccc|cc}
\toprule[1.0pt]
Baseline                  & Evolve          & $M_{Aug}$                 & $E_{Aug}$  & Selector              & $\text{SR}_{1}$    & $\text{SR}_{4}$  \\ \hline 
\checkmark &                           &                           &                           &                      & 63.4\%    & 63.4\%                  \\
\checkmark & \checkmark &                           &                           &                      & 67.6\%     & 67.4\%                \\
\checkmark & \checkmark &\checkmark  &                           &                        & 67.3\%   & 67.2\%              \\
\checkmark & \checkmark &  & \checkmark                          &                         & -   & 73.4\%              \\
\checkmark & \checkmark & \checkmark & \checkmark &                        & 69.6\%  & 74.6\%                    \\
\checkmark & \checkmark & \checkmark & \checkmark & \checkmark & 69.8\% & \textbf{75.4\%}           \\
\bottomrule[1.0pt]
\end{tabular}
\vspace{-0.4cm}
\end{table}
\subsubsection{Ablation study on different components}\label{EXP:comp}
We analyze the impact of SEIL’s components in Table \ref{tab:components}, starting from the 8-shot baseline. We first discuss the success rate after four round evolution. Training with only baseline-recorded demonstrations achieves a higher SR (67.4\%) than using both baseline and $M_{Aug}$-recorded data (67.2\%). This drop is attributed to inconsistent trajectories from $M_{Aug}$ under the same environment $E$, where identical observations lead to different actions - undesirable for imitation learning. To be more specific, given an initial observation $O_{0}$, the Base model and $M_{Aug}$ produce different actions, $A_{0}^{Base}$ and $A_{0}^{M_{Aug}}$, respectively. This discrepancy introduces inconsistency into the evolutionary process. Introducing $E_{Aug}$ allows $M_{Aug}$ to operate from varied initial states, mitigating this inconsistency and improving SR to 74.6\%. Finally, applying our selector to retain only the most diverse and informative samples further boosts SR to \textbf{75.4\%}. We further study the importance of $E_{Aug}$ by analyzing the success rate after the first and fourth rounds. In the absence of $E_{Aug}$, the model is affected by the inconsistency between different supervisions, which results in a performance decline from 67.6\% to 67.4\%. With environment-level augmentation, the model achieves a 73.4\% success rate at the fourth round, reflecting a 7\% improvement compared to a single-round evolution (67.6\%).


\begin{figure}[h]
    \centering
    \setlength{\abovecaptionskip}{-0.6cm}
    \includegraphics[width=0.5\textwidth]{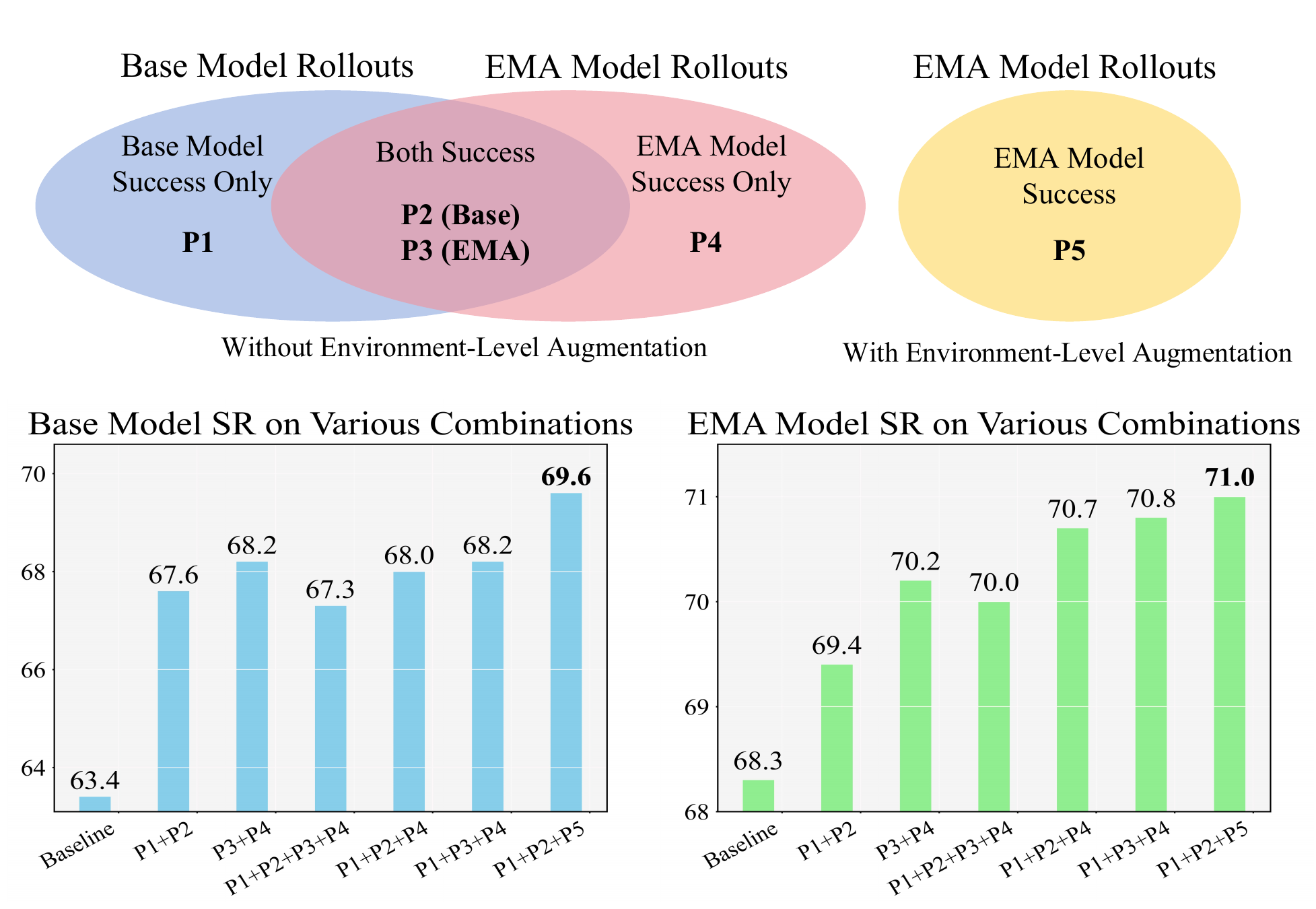}
    \caption{Ablation study on evolution with different types of rollouts. We first record rollouts using the Base model and EMA model under identical environment settings. We divide the demonstration pool into three categories: Base Model success only ($\text{P1}$), EMA model success only ($\text{P4}$), and cases where both succeed ($\text{P2}$, $\text{P3}$). We further extend this setup by recording EMA model under varied environment settings, denoted as $\text{P5}$. We report the Success Rate (SR) of both Base and EMA model evolved on various rollouts combinations.}
    \label{fig:demo}
    \vspace{-0.3cm}
\end{figure}
\subsubsection{Effects of various rollouts combinations}
Figure \ref{fig:demo} reports the impact of various rollouts combinations generated by different models after a single round of evolution, providing further insights into the effect of proposed dual-level augmentation. We first analyze the individual rollout pools generated by the Base model ($\text{P1}+\text{P2}$) and the EMA model ($\text{P3}+\text{P4}$). Compared to Base model rollouts, which provide a 4.2\% improvement over the baseline, evolving on EMA model rollouts further improves the few-shot Base model by 4.8\% and the EMA model by 1.9\%, indicating that EMA rollouts contain more informative demonstrations. Next, we combine both pools ($\text{P1}+\text{P2}+\text{P3}+\text{P4}$) and evolve the model for one round. Despite the larger number of demonstrations for evolving, performance of Base model declines to 67.3\%, lower than evolving on either individual pool. To investigate this, we introduce two complementary groups: adding only EMA-successful demonstrations to the Base pool ($\text{P1}+\text{P2}+\text{P4}$) and vice versa ($\text{P1}+\text{P3}+\text{P4}$). Augmenting the Base pool with $\text{P4}$ increases the success rate to 68.0\% (vs. 67.6\% on $\text{P1}+\text{P2}$), demonstrating that additional diverse demonstrations facilitate model evolution. Conversely, incorporating Base-successful rollouts into the EMA-generated pool ($\text{P1}+\text{P3}+\text{P4}$) results in an identical success rate of 68.2\%. We investigate this by analyzing the number of supplementary demonstrations, which reveals that, since the EMA model outperforms the Base model (68.3\% vs. 63.4\%), the $\text{P1}$ set contributes only about one demonstration per task on average, limiting its impact. Finally, by introducing environment-level augmentation ($E_{Aug}$), the supplemented pool becomes a more diverse dataset. The union of dual-level augmentation and Base model generated rollouts ($\text{P1}+\text{P2}+\text{P5}$) then achieves the best performance of \textbf{69.6\%}, underscoring the effectiveness of our proposed dual-level augmentation.

\begin{table*}[t]
\centering
\caption{Ablation of number of rollouts per round on simulator.}
\label{tab:num_attempts}
\begin{tabular}{c|c|ccccc}
\toprule[1.0pt]
Rollouts\textbackslash{}Round & Baseline                & First Round & Second Round & Third Round & Fourth Round & Fifth Round \\ \hline
10                            & 63.4\% & 65.1\%      & 68.4\%       & 68.6\%      & 70.8\%       & 72.0\%      \\ \hline
20                            &  63.4\%                       & 67.8\%      & 71.8\%       & 72.4\%      & 73.2\%      & \textbf{74.2\%}      \\ \hline
50                            &  63.4\%                       & 70.6\%      & 72.4\%       & 71.4\%      & -            & -           \\ \hline
100                           & 63.4\%                        & 72.1\%  & 71.7\%           & -           & -            & -           \\ \bottomrule[1.0pt]
\end{tabular}
\vspace{-0.2cm}
\end{table*}

\subsubsection{Effect of the attempt numbers}
Table \ref{tab:num_attempts} presents the ablation study on the number of recorded demonstrations (RD) used per round. In this setting, we apply $E_{Aug}$ to the base model and evolve with all demonstrations without selection. The demonstration pool is incrementally expanded as new trajectories are generated. As shown in Table \ref{tab:num_attempts}, when adopting only 10 rollouts per round, the evolution process becomes ineffective, as the model exhibits only marginal gains with each new demonstration—for instance, the success rate improves by merely 0.2\% from the second to the third round. However, when the number of rollouts increases to 50 or 100, the model performance quickly saturates in the early rounds. For instance, in the 100-rollouts setting, the success rate even drops to 71.7\% in the second round, which we attribute to overfitting. 
With a large number of demonstrations collected in a single round, the model tends to overfit by rapidly adapting to the current RD, rather than gradually refine the RD through staged evolution. Compared to other settings, using 20 rollouts per round yields the best results, reaching a peak success rate of 74.2\% at the fifth round.

\begin{table}[t!]
\centering
\caption{Ablation Study on Selection Scheme. Mixed represent half demonstrations in descending order and the rest half in ascending order.}
\label{tab:ablation_selection}
\begin{tabular}{c|ccccc}
\toprule[1.0pt]
Scheme  & Uniform & Descending & Ascending & Mixed  \\ \hline
SR  & 70.8\%  & 72.0\%    & \textbf{72.7\%}    & 72.2\% \\ \bottomrule[1.0pt]
\end{tabular}
\end{table}

\begin{table}[t]
\centering
\caption{Accuracy of pretrained classification task.}
\label{tab:select_per}
\begin{tabular}{c|cc}
\toprule[1.0pt]
Shot\textbackslash{}Input & Sequence & Img+Sequence \\ \hline
1 shot                    & 53.0\%   & 57.0\%       \\
2 shot                    & 56.0\%   & 61.0\%       \\
4 shot                    & 54.5\%   & 63.5\%       \\
8 shot                    & 74.5\%   & 85.0\%     \\ \bottomrule[1.0pt]
\end{tabular}
\vspace{-0.3cm}
\end{table}



\subsubsection{Effect of the selector}
\paragraph{Effect of the selection scheme}
Table \ref{tab:ablation_selection} reports the success rate under different selection strategies.  We select 20 demonstrations per task from a 50-attempts Base and $M_{Aug}$ model generated demonstration pool. Compared to random sampling, all selector-based approaches yield consistently higher performance. 
Notably, selecting demonstrations with the lowest confidence scores achieves the best performance (72.7\%), highlighting the benefit of prioritizing diversity. 
After pre-training the selector on task classification, it effectively identifies the task associated with each demonstration.
Demonstrations with low confidence scores are those that deviate most from the few-shot expert demonstrations, offering complementary diversity that enhances policy training. 
Additionally, the Mixed strategy outperforms the Descending strategy (72.2\% \textit{vs.} 72.0\%), further supporting the importance of including diverse rather than only high-confidence samples.
\paragraph{Effect of the trajectory classification task}
Table \ref{tab:select_per} presents the classification accuracy of the selector on the validation set. When comparing different input modalities, we observe that incorporating images enables the selector to classify trajectories more precisely, leading to improved average 7\% accuracy in most cases. However, even without the images, the selector demonstrates competitive classification performance, indicating that the trajectories themselves carry task-specific categorical attributes. The experimental results also indicate that the selector possesses the ability to understand the trajectory scenes.

\section{Limitations}
Despite SEIL’s effectiveness in few-shot settings, several limitations remain. First, the framework relies on a simulator to generate demonstrations, which may not be accessible in real-world scenarios. However, compared to the tedious process of manual data collection, our method offers an alternative that can efficiently scale datasets in a more automated and cost-effective manner. Second, SEIL requires multi-stage training, which introduces additional evolving time compared to one-stage approaches. 

\section{Conclusion}
In this paper, we proposed Self-Evolved Imitation Learning (SEIL), a novel framework that progressively evolves a weak few-shot policy through interactions with a simulator. 
SEIL introduces a dual-level augmentation strategy to generate diverse demonstrations and a lightweight trajectory selector to identify informative samples for efficient policy refinement. 
Extensive experiments across multiple benchmarks validate the effectiveness of SEIL and demonstrate the individual contributions of its core components.

\bibliographystyle{IEEEtran}
\bibliography{calib}

\end{document}